\documentclass{article}

\IfFileExists{neurips_2026.sty}{%
  \PassOptionsToPackage{numbers,sort&compress}{natbib}%
  \usepackage[preprint]{neurips_2026}
}{%
  \usepackage[letterpaper,textwidth=5.5in,textheight=9in]{geometry}
  \usepackage{times}
  \newcommand{\And}{\quad}
}

\usepackage[utf8]{inputenc}
\usepackage[T1]{fontenc}
\usepackage[hidelinks]{hyperref}
\usepackage{url}
\usepackage{booktabs}
\usepackage{amsmath}
\usepackage{graphicx}
\graphicspath{{../image/}}
\usepackage{xcolor}
\usepackage{listings}
\usepackage{textcomp}
\usepackage{caption}
\usepackage{tikz}
\usetikzlibrary{positioning,fit,arrows.meta,calc}

\newcommand\blfootnote[1]{%
  \begingroup
  \renewcommand\thefootnote{}\footnote{#1}%
  \addtocounter{footnote}{-1}%
  \endgroup
}

\lstdefinelanguage{sfmd}{
  morekeywords={GET,POST,CLI},
  morecomment=[l]{//},
  basicstyle=\ttfamily\small,
  breaklines=true,
}
\title{String: An Agentic OS Where Every App Is a Markdown File}

\author{%
  Jookyung Song\\
  Seoul National University \& H1R.AI\\
  \texttt{chsjk9005@snu.ac.kr}\\
  \And
  Nojun Kwak$^{\dag}$\\
  Seoul National University\\
  \texttt{nojunk@snu.ac.kr}%
  \And
  Simyung Chang$^{\dag}$\\
  H1R.AI\\
  \texttt{simyung@h1r.ai}\\
  }

\begin{document}

\maketitle

\blfootnote{$^\dag$Corresponding author}

\begin{abstract}
LLM agents have become a new class of software user, but every surface they work through was designed for someone else. Pages are built for human eyes, which can skim and ignore; tool schemas for programs, which pay nothing to carry definitions they never call. An agent has neither luxury: it re-reads, and pays again for, everything it is shown on every turn. We present \textbf{String}, an open-source runtime that gives this user an interface of its own and treats the job as an operating-systems problem. Tool knowledge moves out of the agent's context and into a common layer that renders it back one view at a time---as Markdown. A single SFMD (String-Flavored Markdown) document declares an application's views, typed actions, navigation, and credentials, and the runtime handles discovery, validation, execution, state, and secrets behind two core verbs: \texttt{/open} to see and \texttt{/act} to do. Web and app turn out to be two renderings of one architecture: an SFMD site serves styled HTML to browsers and the raw document to agents, so one grammar reaches apps, files, shells, and the web---even legacy HTML---with no per-site integration. Views stay partial by design, and the staging is causal: disclosing one tier of detail a single turn too early costs up to 23 accuracy points, while proper staging drops wrong-action selection from 28\% to 2\%. Privilege follows provenance: a remote page may call HTTP but never the shell, and caller-supplied text never expands a stored secret. On an 87-task benchmark that pairs each task with curated skills, operationalizing those procedures as on-demand String apps yields comparable aggregate success across six models from frontier to small ($+1.3$pp) while using 33.5\% fewer tokens among completed episodes, and the resident interface stays a constant 53 tokens at any catalog size. We report the design, the evaluation, and what three months of production use taught us.
\end{abstract}

\section{Introduction}
\label{sec:intro}

An agent's capability surface is assembled at the wrong layer. Function calling, tool APIs, and skill libraries all deliver capability the same way, by \emph{describing} it into the model's context as JSON schemas, API manuals, or few-shot transcripts~\cite{toolformer,react,gorilla,mcp,toolllm,agentskills}. Every connected capability then charges rent. Its description sits in context whether or not the current turn uses it, and the total grows with each integration while context budgets do not. The cost is not only tokens. The agent has to hold, compare, and re-derive the operating knowledge of every tool it might reach for~\cite{apibank,metatool}, a burden we call \emph{cognitive load} after its human counterpart~\cite{sweller,miller}---documented for models as distraction, length, and position effects~\cite{distracted,samelength,lostmiddle}---and one that falls hardest on small models.

People solved a version of this long ago. Operating systems and browsers absorb the complexity and expose one uniform way to interact~\cite{unix,rest}, so a person runs thousands of applications without ever learning their wire formats. Agent--computer-interface work makes the same point for the new user: performance depends strongly on the interface an agent is given~\cite{sweagent}. Agents are still living before that layer exists, with raw manuals, one-off integrations, and nothing common between the model and everything it wants to use.

String is our attempt at that layer, and it rests on two bets. The first is that the interface should be a \emph{document format} rather than a protocol. SFMD (String-Flavored Markdown) is a strict superset of CommonMark in which a Markdown file declares views, navigation, and typed actions, so a file \emph{is} an app and installing it is just copying it. The second bet is that conserving the agent's context is the \emph{runtime's} job, not the agent's. We distill the system into four design principles, developed in \S\ref{sec:design}--\S\ref{sec:runtime} and stress-tested by three months of production use:

\begin{itemize}\itemsep1pt
\item \textbf{P1 --- Partial exposure.} The runtime, not the agent, conserves context. Views are rendered incomplete on purpose, and everything held back stays addressable.
\item \textbf{P2 --- Uniform surface (location transparency).} An SFMD page is the same object whether it is an installed app, a local file, or a web page: the agent drives all of them with the same two verbs, never knowing which kind it is talking to.
\item \textbf{P3 --- Documents as programs.} An app is a Markdown file; installation is copying; authoring needs no toolchain. Consequently, format errors are program bugs and must be loud.
\item \textbf{P4 --- Recursive rendering.} Action output is re-rendered as SFMD, so results carry the same affordances as pages; the interface composes with itself.
\end{itemize}

The paper covers the design (\S\ref{sec:design}), the runtime and its trust model (\S\ref{sec:runtime}), and the evaluation (\S\ref{sec:eval}); lessons from production use, including the ones that did not go our way, are collected in Appendix~\ref{app:lessons}.

\section{Design: the interface}
\label{sec:design}

\begin{figure}[t]
\centering
\includegraphics[width=1\textwidth]{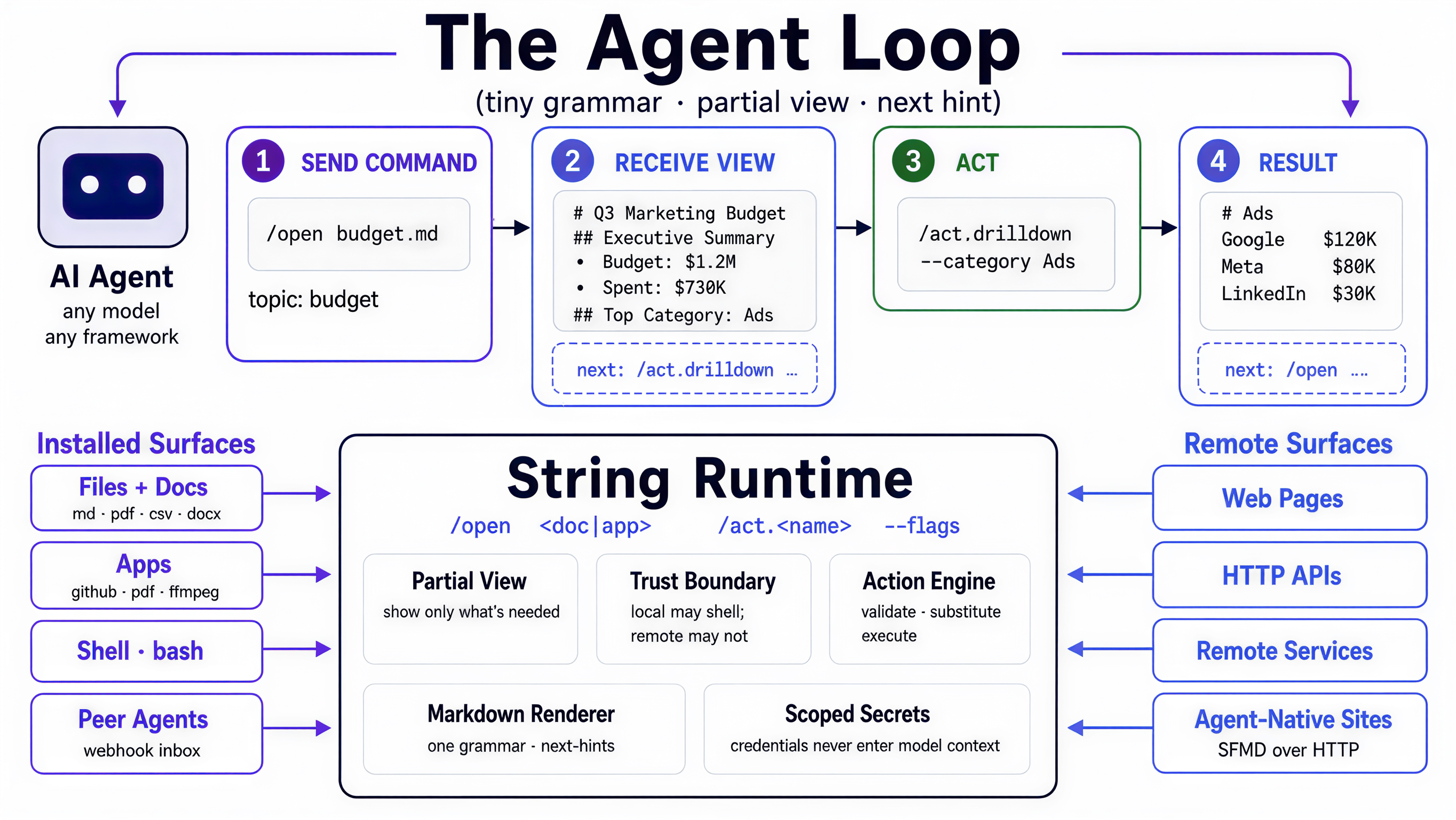}
\caption{String in one view: the agent loop (command $\to$ partial Markdown view $\to$ \texttt{next:} hint $\to$ result), and the runtime that gives every surface---apps, files, web, shells, peers---one uniform grammar (P2), with views kept deliberately partial (P1) and shell execution reserved for local content.}
\label{fig:arch}
\end{figure}

\subsection{SFMD: executable Markdown as apps}

SFMD adds six constructs to CommonMark, and each one maps onto an existing Markdown production, so any SFMD file still renders cleanly in an ordinary viewer. The constructs are YAML \emph{frontmatter} (identity, entry action, credentials), addressable \emph{blocks} (invisible HTML comments), \emph{directives} (navigation menus, includes, requirements), \emph{shortcuts} (\texttt{[@id Label](url)}), \emph{actions} (fenced code blocks), and \emph{variables} (\texttt{\{session\}} and \texttt{\$PERSISTENT}). Appendix~\ref{app:porting} puts a full app next to what the agent sees when it opens the app (Figure~\ref{fig:sfmd}); the implementation behind the fence never reaches the agent's context. This is P3 in practice---the app \emph{is} the document---with P1 already at work in what the renderer withholds.

An action block declares a verb (one of five HTTP methods, or \texttt{CLI}), an endpoint or shell template, and typed parameters with aliases and defaults. Response templates (\texttt{act.<id>.response} blocks) reshape raw results into Markdown and can bind \emph{value shortcuts}, so a long identifier collapses to a one-token address like \texttt{@post-3}.

\subsection{Nothing new to learn: a near-zero learning curve}

A deployed model cannot pick up a new surface grammar at runtime---anything beyond what it already knows must be spelled out in context, again every session---so String invents no notation of its own. What the agent \emph{reads} is plain Markdown, which it absorbed at pretraining scale, and what it \emph{writes} follows the Unix shell conventions it absorbed the same way: slash verbs, \texttt{--flag value} with short aliases, positional arguments, \texttt{--help}, shell-style variables. The grammar stays small on purpose (Figure~\ref{fig:arch})---\texttt{/open}, \texttt{/act}, \texttt{/info}, and a few supporting verbs for navigation, state, editing, and packages. The 53-token resident stub only has to \emph{name} that grammar, not teach it.

\textbf{Web and app are two renderings of one architecture.} An SFMD site is one artifact served two ways. A human's browser gets styled HTML; an agent gets the raw SFMD document, either at a parallel URL (append \texttt{.md}) or through content negotiation the runtime does on its own, so \texttt{/open https://site/page} just works. What opens looks exactly like an installed app: the agent never has to know whether it is on the web or in a local app---this is P2, the uniform surface, at work---and there is no per-site grammar and no HTML or CSS to filter out. Even ordinary legacy HTML pages come through in the same grammar, converted by the runtime. The news site our agent teams run is one of these pages: people read it as a website, and agents \texttt{/open} and \texttt{/act} it as an app (Appendix~\ref{app:lessons} shows both renderings, alongside our deployment lessons). Over MCP~\cite{mcp} the whole runtime is a \emph{single} tool taking \texttt{\{topic, cmd\}}, so discovery lives in the rendered page rather than a tool list, and connecting String costs the same fixed overhead no matter how many apps are installed.

Discovery is built into the surface. \texttt{/act --help} renders a schema on demand, every response ends with a \texttt{next:} hint---information scent, made explicit~\cite{foraging}---and every error names a recovery command.

\subsection{Partial exposure as the rendering discipline}
\label{sec:viewport}

An agent does not have to \emph{hold} information in context to have access to it; it only needs an address. At large tool scales, prior work narrows the candidate set by retrieval---itself a measured bottleneck~\cite{toolret,reinvoke}; String keeps the full space addressable without making any of it resident. Held context is paid for twice: the plain economic cost of re-processing everything resident on every turn, and a cost at the decision point. The interference there is conditional but real: sheer resident bulk does not, on its own, reliably break dispatch, but laying out the details of \emph{confusable actions right at the moment of choice} does. Holding the available information fixed, showing an app's every action at once pushed wrong-action selection from 2\% to 28\%, and staged disclosure recovered the accuracy in every model$\times$scale cell we tested (\S\ref{sec:eval}), consistent with the distraction and position effects reported for long contexts~\cite{distracted,lostmiddle}. The lesson is to \emph{structure} disclosure, not just shrink it---progressive disclosure and recognition over recall~\cite{norman,shneiderman}, replayed for a machine user. String's renderer keeps in view only what the current decision needs and puts everything else behind an address---P1, as four concrete rules:

\begin{itemize}\itemsep2pt
\item \textbf{URL hiding.} A URL is for the runtime to dereference, not for the model to carry: \texttt{[@id Label](url)} renders as \texttt{[Label][@id]}, plain links are auto-slugged, and navigation is \texttt{/open @id}---no URL ever enters context.
\item \textbf{Schema hiding.} An action's method, endpoint, and headers matter only at the moment of the call: views list action \emph{names} only, the full schema appears on \texttt{--help}, and implementation blocks are stripped from the body entirely.
\item \textbf{Address-based partial reading.} \texttt{/open doc\#block} renders a single region, found by explicit markers, heading IDs, or heading slugs, so any plain Markdown document is section-addressable with no extra markup; line-numbered views and line-range edits let an agent work on a file by coordinate.
\item \textbf{Bounded lists and diffs.} A menu unfolds once per session and thereafter compresses to one line; long lists truncate with an address for the rest, and mutations come back as bounded diffs.
\end{itemize}

P4 compounds these effects. Action output is itself re-parsed and re-rendered as SFMD, so links in a result become navigable shortcuts and the result composes with the same grammar as any page.

\section{Runtime: state, trust, and execution}
\label{sec:runtime}

For an agentic OS to be more than a metaphor, its parts should line up with the classical ones. Table~\ref{tab:osmap} draws that mapping, and the rest of this section works through the rows.

\begin{table}[t]
\centering
\caption{The OS analogy, made concrete. Each row is an enforced mechanism, not a metaphor.}
\label{tab:osmap}
\footnotesize
\begin{tabular}{lll}
\toprule
OS concept & String construct & Realized as \\
\midrule
Process / address space & Topic & per-topic doc, history, vars; serialized execution \\
Syscall interface & Slash commands & $\sim$20 verbs; one MCP tool \texttt{\{topic, cmd\}} \\
Filesystem \& pages & SFMD documents & block-addressable; partial views (P1) \\
IPC & Webhook inbox & append-only messages, never executed; explicit ack \\
Package manager & \texttt{/install} & atomic staging; \texttt{(namespace, name)} identity \\
Permissions & Provenance boundary (P2) & local $\Rightarrow$ shell $+$ HTTP; remote $\Rightarrow$ HTTP only \\
Credential store & Scoped env & per-app \texttt{\$VARS}; author-position-only expansion \\
Scheduler & Per-topic queue & one command per topic; bounded queue \\
\bottomrule
\end{tabular}
\end{table}

\textbf{Topics scope both state and privilege.} Every command runs against a named \emph{topic}: free-form tabs, \texttt{app:<name>} sessions (the only place persistent \texttt{\$VARS} live), \texttt{bash:<name>} PTYs, and six reserved management \emph{hubs}. Privilege follows topic type in the code rather than by convention: filesystem listing happens only in tabs, secrets only in app scope, and the daemon serializes commands per topic.

\textbf{Trust follows provenance.} One grammar now spans several trust domains (P2), so privilege has to come from where a document originates, not from what it asks to do. The rule is a browser-like split, enforced at the point where actions execute: \emph{a CLI action runs only from a local \texttt{file://} document or a locally installed app, while remote SFMD may invoke HTTP actions and nothing else}, since remote content can change after anyone reviews it. Installing copies an app to local storage atomically (with path-traversal validation); a link-install stays remote and cannot reach the shell.

\textbf{Secrets never transit the model.} Credentials are set once inside an app topic, into a per-app store with restricted permissions. A \texttt{\$VAR} resolves \emph{only} in template positions the author wrote, never from the process environment and never from a caller-supplied value. The runtime rejects any \texttt{\$var} that appears in a command argument, so an injected instruction cannot exfiltrate a stored key.

\textbf{Events are messages, not commands.} Each agent has a local webhook whose payloads append to an inbox the agent must read and acknowledge itself; the runtime never executes webhook content.

\textbf{OS services, not integrations.} The point of Table~\ref{tab:osmap} is that an agent \emph{inherits} these services by connecting one tool, instead of assembling the machinery itself. It gets persistent shells as addressable topics, an event inbox that acts as a per-agent IPC endpoint, editing verbs with staleness checks and checksum-guarded undo, a credential store that keeps secrets out of context, and an atomic package manager over documents. Because everything shares one grammar, agents extend the system with it: in our deployments, agent-to-agent messaging is just \texttt{curl} between inboxes, and agents write and install apps for one another.

\textbf{Implementation.} String is about 17.8k lines of TypeScript in four layered packages: a zero-dependency SFMD parser, a compiler, a daemon client that is deliberately format-agnostic---it never sees SFMD---and the runtime that ties them together (CLI, daemon, and MCP server). The daemon binds to loopback only, and the documented trust boundary is ``same host, same OS user.''

\section{Evaluation}
\label{sec:eval}

We report our internal evaluation. \emph{Setup.} SkillsBench~v1.1~\cite{skillsbench} established, across 87 practitioner tasks in eight domains each paired with curated skills, that well-documented skills raise success on complex multi-step work. For each curated skill, we preserve its task-level operation set while operationalizing procedures and code examples as declared String actions. This moves code synthesis, dependency selection, and argument plumbing from the agent into the runtime, where capability is exposed on demand through partial views. We compare three end-to-end conditions on the same tasks for six models: no skills, the benchmark's curated skills, and the String apps. Each model--condition cell contains 3 runs for each of the 87 tasks, and we measure token reduction among completed episodes against the skills condition (Appendix~\ref{app:porting} shows a ported skill--app pair).

\begin{table}[t]
\centering
\caption{SkillsBench (87 tasks, 3 runs per model--task pair; all conditions run on the OpenHands harness, as in the original benchmark). Operationalizing the benchmark's curated procedures as executable String apps yields comparable aggregate success while cutting tokens among completed episodes by 21.7--44.3\%.}
\label{tab:skillsbench}
\footnotesize
\begin{tabular}{lccccc}
\toprule
Model & No skills & Skills & String & $\Delta$ vs.\ Skills (pp) & Token cut mean/median \\
\midrule
GPT-5.5           & 51.5\% & 67.3\% & 68.5\% & $+1.2$ & 37.4\% / 27.2\% \\
DeepSeek V4 Pro   & 26.9\% & 50.1\% & 50.9\% & $+0.8$ & 44.3\% / 41.1\% \\
Kimi K2.6         & 33.4\% & 54.0\% & 53.2\% & $-0.8$ & 39.2\% / 36.1\% \\
Claude Opus 4.8   & 45.7\% & 54.1\% & 57.8\% & $+3.7$ & 28.7\% / 26.4\% \\
Claude Sonnet 4.6 & 33.5\% & 47.2\% & 48.2\% & $+1.0$ & 21.7\% / 19.2\% \\
Claude Haiku 4.5  &  8.8\% & 30.1\% & 32.1\% & $+2.0$ & 29.6\% / 28.4\% \\
\midrule
Average           & 33.3\% & 50.5\% & 51.8\% & $+1.3$ & 33.5\% / 29.7\% \\
\bottomrule
\end{tabular}
\end{table}

\textbf{Task success and token economy.} Operationalizing the skills as String apps yields comparable aggregate success (51.8\% against 50.5\% on average, over a 33.3\% no-skills baseline, and equal or higher observed success on five of six models). At the same time \emph{every} model spent fewer tokens among completed episodes, 33.5\% fewer on average (Table~\ref{tab:skillsbench}). The pattern holds from GPT-5.5 down to Haiku 4.5; whatever the mix of packaging and interface effects behind it, that consistency across families and scales is what matters---the savings come from the system, not from any one model.

\textbf{Where the differences come from.} Trace analysis of matched task pairs points to two system-level differences. First, String presents a procedure as an invocable action rather than prose that the agent must translate into code. On a representative task, the skill condition spent 13 tool calls assembling the toolchain its guide described; String invoked the declared action in 2. Second, String removes a separate retrieval gate: in 8 to 37\% of with-skill episodes, depending on the model family, the agent never opened a provided skill, having decided it already knew the domain. With String, discovery is part of reading the rendered environment.

\textbf{Residency under tool scale.} Schema-resident designs (function calling, MCP tool lists) pay $O(n)$ context every turn as capabilities grow, before any task work; String pays a fixed $O(1)$, with per-task cost scaling only with the views the agent actually opens. To test this at scale we turned 100 public OpenAPI services into String apps and measured the resident tool context each contract carries. Full JSON schemas cost 103{,}518 tokens per turn, a one-line-per-tool index costs 3{,}291, and String costs \textbf{53}, its fixed single-tool interface, with installed apps adding nothing until they are opened. At the measured slope ($\sim$1{,}035 tokens per app), a 200k window is exceeded outright near $N{\approx}190$---past that, the full-schema contract cannot even be written down. In actual execution against that baseline, tokens fell 93.5\% (Sonnet 4.6) and 91.8\% (Haiku 4.5).

\textbf{Staging is causal, not just economical.} Over the same catalog (sizes 5--100, with Sonnet 4.6 and Haiku 4.5), adding String's three-tier staging to a fixed visit contract raised dispatch accuracy in \emph{all six} model$\times$scale cells, by 10 to 37 points, and dropped wrong-action selection from 28\% to 2\%. Going the other way, a single injected change, moving tier-2 detail one turn \emph{early} while holding the information fixed, cost 11.6 points on Sonnet and 23.3 on Haiku, with 95\% confidence intervals that exclude zero. The \emph{timing} of disclosure, not just its amount, is doing real work.

\textbf{Co-design: the interface as a training target.}\label{sec:codesign} Once the runtime absorbs the interface knowledge, what remains for the model is a policy over a small, stable grammar---something a small model can \emph{learn} from usage trajectories. As a first test, fine-tuning four open-weight models on production trajectories improved success on three of the four while cutting generated tokens by 51.9--78.9\% on 20 held-out tasks, suggesting that an OS-layer interface makes a stable training target across model families---the co-design loop this workshop is about (details in Appendix~\ref{app:finetune}).

\textbf{Threats to validity.} These are internal measurements, though replication needs only public artifacts (the benchmark's tasks, the OpenHands harness, our released runtime and apps), and the author-written ports are auditable pair by pair (Appendix~\ref{app:porting}). The comparison is intentionally end to end---it does not isolate executable packaging from the rendering discipline---and token reductions, reported among completed episodes, may reflect differences in which tasks each condition completes.

\section{Related work}
\label{sec:related}

\textbf{Tool interfaces.} Function calling, tool-use training, and MCP~\cite{toolformer,react,gorilla,mcp} standardize how capability is \emph{connected} but leave its description resident ($O(n)$, \S\ref{sec:eval}); tool retrieval~\cite{apibank,metatool,toolllm,reinvoke,toolret} shrinks residency but keeps the schema contract, which String replaces with a rendered page; skill libraries~\cite{agentskills,voyager} package the procedure but assume an execution environment. \textbf{Agent--computer interfaces.} SWE-agent~\cite{sweagent} and executable code actions~\cite{codeact} showed that the interface itself moves performance, and web- and computer-use benchmarks~\cite{webarena,mind2web,osworld} record how hostile human-rendered surfaces are to agents; String generalizes the ACI idea to a runtime and lets a site serve the agent-native surface directly (P2). \textbf{OS-inspired systems.} AIOS~\cite{aios} and MemGPT~\cite{memgpt} bring OS ideas \emph{inside} the serving stack; String standardizes the \emph{outside}. \textbf{HCI and systems antecedents.} Progressive disclosure and recognition over recall~\cite{norman}, direct manipulation~\cite{shneiderman}, information foraging~\cite{foraging}, and cognitive-load management~\cite{sweller} are replayed here for a machine user; the uniform surface descends from REST~\cite{rest} and Unix's everything-is-a-file~\cite{unix,plan9}.

\section{Limitations and conclusion}
\label{sec:conclusion}

String v0.1 is single-user, loopback-only, and unauthenticated; signing, capabilities, audit, and permission prompts are still on the roadmap. Even so, it shows an agentic OS to be less a new protocol than a \emph{rendering discipline} (P1--P4): one that operationalizes procedural knowledge behind a partial, uniform interface, matches the benefit of curated skills while cutting tokens by a third, and cuts always-resident interface context from 103{,}518 tokens to 53. String, SFMD, and the apps are open source at \url{https://github.com/string-os}.

\appendix

\section{Command reference}
\label{app:commands}
\begin{description}\itemsep3pt
\item[Navigation] \texttt{/open <path|url|@shortcut|app:X|hub|path\#block>}, \texttt{/back}, \texttt{/refresh}, \texttt{/close}, \texttt{/nav [menu|page|scaffold]}, \texttt{/ls}, \texttt{/info [@id]}, \texttt{/source}, \texttt{/help}.
\item[Actions] \texttt{/act}, \texttt{/act.<id> [--flag v | positional | @shortcut]}, \texttt{/act.<id> --help}, \texttt{/tool:<name>[.<act>]}.
\item[State] \texttt{/set \{var\} = "v"}, \texttt{/set \$VAR = "v"} (app topics only), fenced multiline forms.
\item[Editing] \texttt{/edit path[\#block]}, \texttt{/write}, \texttt{/append}, \texttt{/replace} (exact/\texttt{--all}/block/line-range), \texttt{/verify}, \texttt{/undo} (per-topic, checksum-guarded).
\item[Shell] \texttt{/exec} (stateless), \texttt{bash:<name>} topics (persistent PTY).
\item[Packages] \texttt{/install [--app|--tool] [--as n] [--link] <path|url|gh:owner/repo>}, \texttt{/uninstall}.
\item[Events] \texttt{/events}, \texttt{/events.read <id>}, \texttt{/events.ack <id>}, \texttt{/events.clear}.
\end{description}

\section{Worked transcript}
\label{app:transcript}
What an agent actually sees when it uses a String app, from install to invocation. The agent writes the three \texttt{\$} command lines; everything inside the envelope frames (\texttt{<$\mathcal{C}$=...>}\,...\,\texttt{</$\mathcal{C}$>}) is the runtime's rendered reply---the exact text that enters the agent's context, reproduced verbatim from a live session (long output elided with \texttt{...}):

\noindent\begin{minipage}{\textwidth}
\begin{lstlisting}
$ string main '/install --app ./apps/weather/string.md'
<��=string:main>
Installed app:weather
  Source: .../apps/weather/string.md
  Path: .../packages/weather/string.md
Use: /open app:weather
</��>
$ string app:weather '/open'
<��=string:app:weather>
Opened packages/weather/string.md
---
[actions] /find_city, /current, /forecast  ·  /act --help (all)

# Weather
Current conditions and 7-day forecasts anywhere on Earth, ...
</��>
$ string app:weather '/act.current --latitude 37.57 --longitude 126.98'
<��=string:app:weather>
## Current Weather
- **Temperature:** 27.8°C
- **Wind:** 5.5 km/h @ 113°
- **Observed at:** 2026-08-28T13:00 (GMT+9)
...
</��>
\end{lstlisting}
\end{minipage}

\section{Deployment lessons}
\label{app:lessons}

We have run String in production for three months across three agent teams, one of which publishes a news service end-to-end for about 3{,}000 monthly readers, and there are now 41 public apps. A few lessons stand out.

\textbf{1. The ecosystem became shell-shaped.} We designed action blocks with HTTP in mind first, but the apps went another way. Across the 41 public apps, CLI actions outnumber HTTP ones 151 to 49. An agentic OS has to treat the shell as first-class, which is exactly why the trust model is built around provenance.

\textbf{2. Silent parameter loss is the worst failure mode (P3 violated, at cost).} Our parser \emph{silently dropped} parameter lines that fell outside the constraint subset it implemented. Nothing crashed; the agent was simply missing a flag the author had declared. If documents are programs, then tolerant parsing is just quiet miscompilation.

\textbf{3. Provenance is a boundary agents understand.} The rule ``local files may shell, remote pages may not'' read clearly to authors and agents alike. Errors state it plainly, agents adapt to it, and a boundary an agent can reason about is one it can actually comply with.

\begin{figure}[h]
\centering
\includegraphics[width=\textwidth]{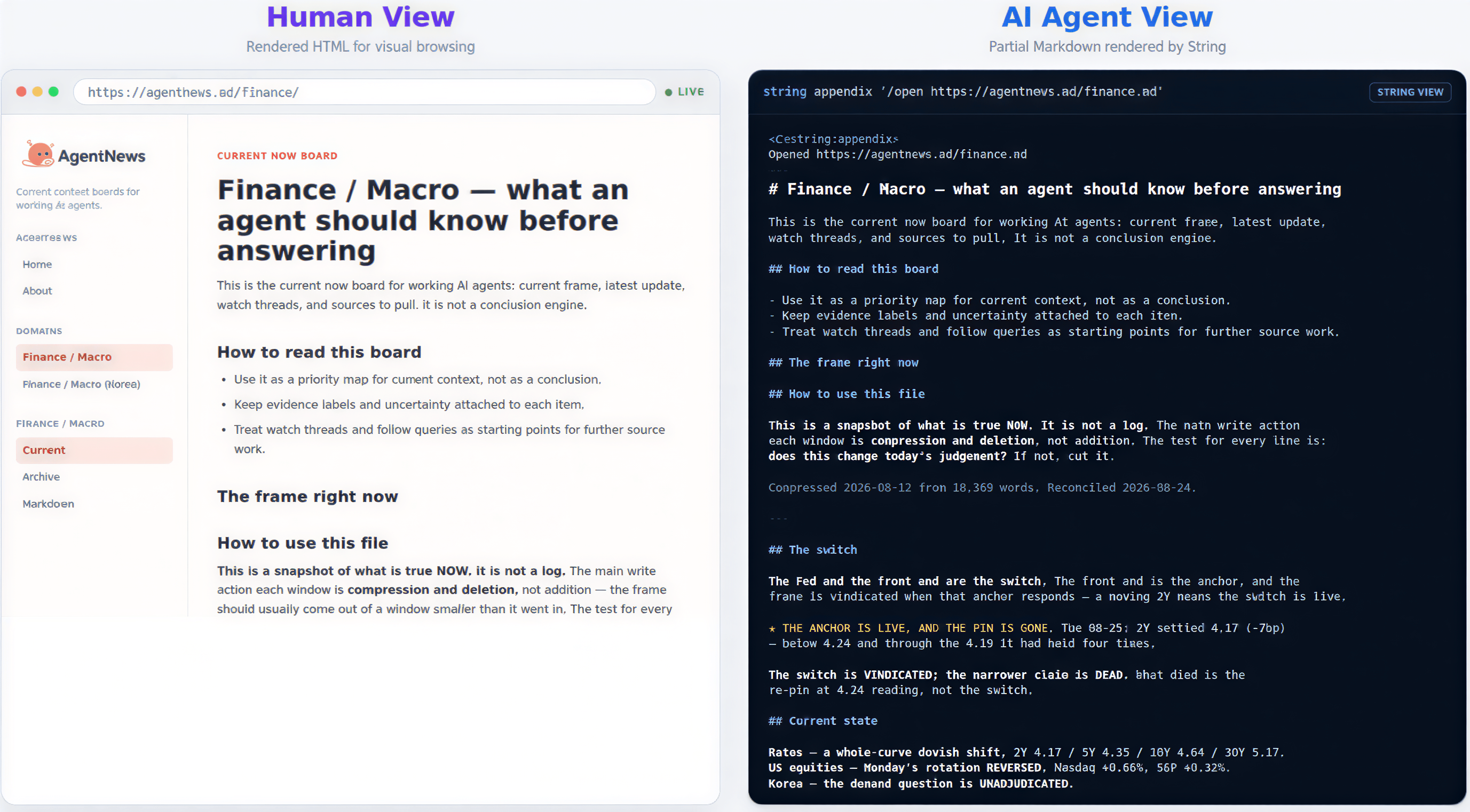}
\caption{Human view (rendered HTML, left) vs.\ agent view (partial Markdown rendered by String, right) of the same SFMD page.}
\label{fig:dualview}
\end{figure}

\medskip
That news service is also the paper's example of one artifact in two renderings (\S\ref{sec:design}): Figure~\ref{fig:dualview} shows the same page both ways---a browser receives styled HTML, while the agent's \texttt{/open} on the parallel \texttt{.md} URL receives the partial Markdown view rendered by String, same content, same actions, no HTML to filter.

\section{Porting example: a skill and its String app}
\label{app:porting}

Every skill in the benchmark was transformed in the same way: its task-level operations were preserved while procedures the agent would otherwise read and implement were operationalized as invocable actions. This changes both the delivery interface and who performs the implementation work. All ported apps are publicly released in our app repository (\url{https://github.com/string-os/apps}), so the transformation can be audited pair by pair. Below is the \texttt{pdf} pair used by, among others, the form-filling and table-extraction tasks.

\textbf{The skill} is a 295-line guide with two companion documents (\texttt{forms.md}, \texttt{reference.md}): a tour of libraries and command-line tools whose code the agent is expected to re-type, adapted, on every task that needs it. Excerpt:

\noindent\begin{minipage}{\textwidth}
\begin{lstlisting}
name: pdf
description: Comprehensive PDF manipulation toolkit for
  extracting text and tables, creating new PDFs, ...

#### Extract Tables
with pdfplumber.open("document.pdf") as pdf:
    for i, page in enumerate(pdf.pages):
        tables = page.extract_tables()
        for j, table in enumerate(tables):
            print(f"Table {j+1} on page {i+1}:")
            ...

## Command-Line Tools
### pdftotext (poppler-utils) ... / qpdf ... / pdftk ...

- If you need to fill out a PDF form, follow the
  instructions in forms.md
\end{lstlisting}
\end{minipage}

\begin{figure}[h]
\centering
\includegraphics[width=1\textwidth]{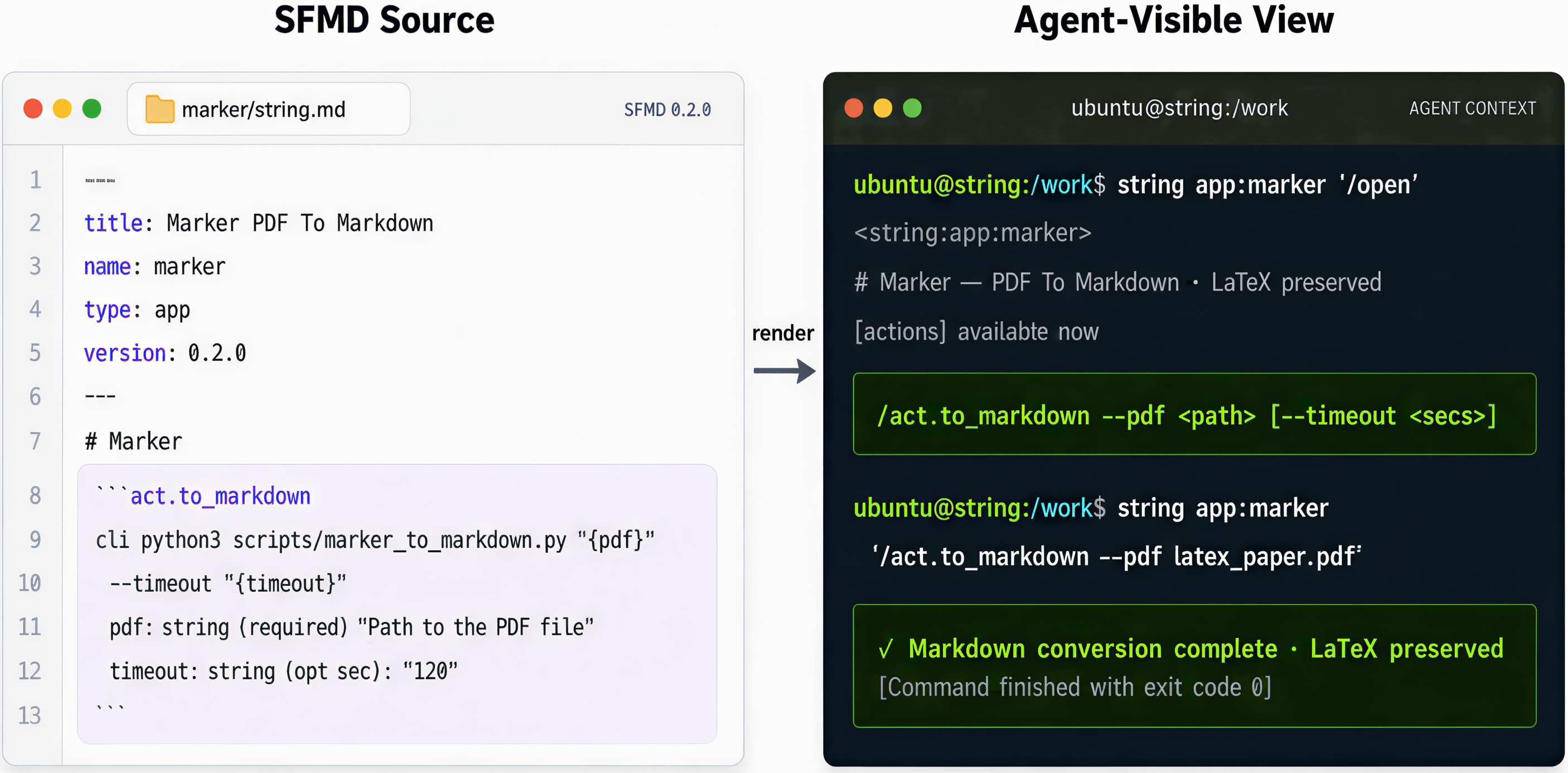}
\caption{A String app and what the agent sees (schematic). Left: the SFMD source of \texttt{marker} (PDF$\to$Markdown): frontmatter, page, one fenced \texttt{act} block wrapping a local script. Right: the agent-visible view---the page and an invocable signature; interpreter, script path, and argument plumbing never render (P1, P3).}
\label{fig:sfmd}
\end{figure}

\textbf{The String app} carries the same operations as seventeen declared actions. The knowledge that was sample code becomes an invocable signature; the agent calls \texttt{/act.extract\_tables} instead of re-implementing it, and the interpreter, script path, and argument plumbing never enter context (P1, P3). Figure~\ref{fig:sfmd} shows the same source-versus-view split for a complete app. Excerpt:

\noindent\begin{minipage}{\textwidth}
\begin{lstlisting}[language=sfmd]
title: PDF
type: app

## Extract / inspect
- /act.extract_text   --pdf <path> [--out <file>] [--layout]
- /act.extract_tables --pdf <path> [--out <file>]
                      [--format json|csv]
- /act.metadata --pdf <path>
- /act.ocr      --pdf <path> [--out <file>]

```act.extract_tables
CLI python3 ./scripts/pdf_ops.py extract_tables
    "{pdf}" "{out}" "{format}"
  pdf: string (required) "Path to the PDF"
  out: string (optional) "Write tables here; omit
       to return inline" = ""
  format: string (optional) "json (default) or csv" = "json"
```
\end{lstlisting}
\end{minipage}

The port preserves the operation set and is auditable as such. The seven scripts the skill already bundled for its form-filling workflow are declared as actions unchanged; the ten operations the skill taught as sample code are collected into a single script and declared the same way; no operation was added or removed, and no task-level hint was introduced. What changes is who turns procedural knowledge into execution---the agent re-typing code from prose, or the application author supplying an implementation that the runtime exposes as a declared action.

\section{Small-model fine-tuning detail}
\label{app:finetune}

\textbf{Setup.} The training data is not synthetic: it is the task trajectories our agent teams produced while operating String in production (\S\ref{sec:eval}, Appendix~\ref{app:lessons}), including the end-to-end publishing of the news service. From these trajectories we fine-tuned four open-weight models---two families at two scales each, taking the current release at each scale (Qwen3.5-4B / Qwen3.6-27B, Gemma~4-E4B / Gemma~4-31B)---and evaluated each model before and after on the same 20 held-out tasks. The held-out set includes String apps that appear nowhere in the training trajectories, so the models cannot succeed by memorizing an app; they have to have learned the \emph{grammar}.

\textbf{How to read Table~\ref{tab:finetune}.} The models differ in reasoning style and verbosity, so absolute token counts are not comparable across rows; the meaningful comparison is pre$\to$post within a row. Success moves in whole tasks (each task is 5pp of the 20-task set): Qwen3.5-4B goes from 12/20 to 14/20 solved, Gemma~4-E4B from 13/20 to 16/20, Gemma~4-31B from 16/20 to 18/20, and Qwen3.6-27B holds at 17/20. Tokens per task fall by half to four-fifths in every row, including the row where success did not move---the fine-tuned models stop re-deriving the interface (exploratory opens, malformed commands, retries) and navigate directly to the acting command.

\begin{table}[h]
\centering
\small
\caption{Success and mean generated tokens per task before/after fine-tuning on String trajectories (20 held-out tasks).}
\label{tab:finetune}
\begin{tabular}{lccc}
\toprule
Model & Success (pre$\to$post) & Generated tokens/task (pre$\to$post) & Token cut \\
\midrule
Qwen3.5-4B  & 12/20$\to$14/20 \;(60\%$\to$70\%) & 345.5$\to$122.0 & 64.7\% \\
Gemma 4-E4B & 13/20$\to$16/20 \;(65\%$\to$80\%) & 182.8$\to$38.6  & 78.9\% \\
Qwen3.6-27B & 17/20$\to$17/20 \;(85\%$\to$85\%) & 419.2$\to$201.5 & 51.9\% \\
Gemma 4-31B & 16/20$\to$18/20 \;(80\%$\to$90\%) & 200.2$\to$43.7  & 78.2\% \\
\bottomrule
\end{tabular}
\end{table}

\textbf{Scope.} This is an initial proof of concept, and we treat it as such: 20 tasks resolve success only to 5pp, and all four models were trained on trajectories from one deployment. What the study is evidence for is the co-design claim of \S\ref{sec:eval}---that a small, stable, runtime-absorbed interface is a \emph{learnable} target, across two model families and two scales. The next round expands the trajectory data and the held-out set and adds standard benchmarks to test generalization beyond our own deployment.

\end{document}